\pdfoutput=1
\documentclass{article}

\usepackage[preprint]{neurips_2023}

\usepackage{booktabs} 
\usepackage{url}             
\usepackage{nicefrac}          
\usepackage{microtype}         
\usepackage{xcolor}            
\usepackage{lmodern}
\usepackage{amsmath}
\usepackage{graphicx}
\usepackage{amsfonts}       
\usepackage{caption}
\usepackage{multirow}
\usepackage{subcaption}
\usepackage{authblk}
\usepackage[bookmarks=false]{hyperref}
\PassOptionsToPackage{[numbers]}{natbib}

\title{The Disagreement Problem in Faithfulness Metrics}

\author[1,*]{Brian Barr}
\author[1]{Noah Fatsi}
\author[1]{Leif Hancox-Li}
\author[2]{Peter Richter}
\author[2]{Daniel Proano}
\author[2]{Caleb Mok}
\affil[1]{AI Foundations, Capital One, Mclean VA}
\affil[2]{Applied ML, Capital One, Boston MA}
\affil[ ]{\textit {\{brian.barr,noah.fatsi,leif.hancox-li,peter.richter,daniel.proano,caleb.mok\}@capitalone.com}}
\affil[*]{corresponding author}

\begin{document}










\maketitle

\begin{abstract}
The field of explainable artificial intelligence (XAI) aims to explain how black-box machine learning models work. Much of the work centers around the holy grail of providing post-hoc feature attributions to any model architecture. While the pace of innovation around novel methods has slowed down, the question remains of how to choose a method, and how to make it fit for purpose.  Recently, efforts around benchmarking XAI methods have suggested metrics for that purpose---but there are many choices.  That bounty of choice still leaves an end user unclear on how to proceed.  This paper focuses on comparing metrics with the aim of measuring faithfulness of local explanations on tabular classification problems---and shows that the current metrics don't agree; leaving users unsure how to choose the most faithful explanations.
\end{abstract}

\section{Introduction}
\label{intro}

XAI is a field that aims to create techniques that explain black-box machine learning models. While there is a growing body of work on mechanistic interpretability \cite{hernandez2022scaling}, which aims to describe the actual mechanisms of model predictions by looking at model components, much of the XAI literature has focused on post-hoc explanations, which aim to create explanations without  depending on the specifics of internal model mechanisms. Within the post-hoc explainability literature, feature attribution methods \cite{lundberg2017unified, ribeiro2016should} have been particularly prominent: methods where the explanation for a particular model prediction is series of numbers representing how important each feature is. Other XAI approaches, like counterfactuals and example-based approaches, provide fundamentally different types of outputs \cite{10.1145/3351095.3372850, kim2014bayesian}.

Feature attribution methods face the challenge of proving that their outputs are faithful to the model's behavior. Various faithfulness metrics have been proposed, some of them as part of XAI benchmarks~\cite{JMLR:v24:22-0142,Chirag_2022}. However, it is unclear how well these different metrics correlate with one another, or what use cases each metric is suitable for. In this paper we ask the meta-question of deciding which measures of faithfulness work well: we benchmark faithfulness metrics.

Feature attribution methods also face the challenge of different methods disagreeing with one another \cite{krishna2022disagreement}. Inspired by this finding, we extend it to look at disagreements between evaluation metrics for XAI. We appropriate two recently introduced methods of evaluating explanations---ablation~\cite{hameed2022based} and topological data analysis~\cite{xenopoulos22}---to tackle the problem of evaluating XAI metrics. We then compare them to other XAI metrics on a variety of different explanation methods, baselines for those explanation methods, and tabular datasets. 


Our paper points to a gap between theory and practice: We have many faithfulness metrics, but they do not correlate well with one another. Practitioners evaluating explanations on faithfulness lack guidance on which faithfulness metric they should use. It may be that similarly to how one selects different accuracy metrics based on the application context, faithfulness metric selection is also highly contextual. If so, more work needs to be done to figure out which faithfulness metrics are suitable for which contexts.

\section{Previous work}

\subsection{Post-hoc explainability challenges}

Post-hoc explainability methods face the challenge of determining whether their outputs are good. This challenge is complicated because, to begin with, there are different ways in which we can define ``goodness''. Stability, faithfulness to the original model, and fairness are just some of the desiderata identified for XAI outputs so far~\cite{Chirag_2022}. In addition, user studies have found that XAI methods may not necessarily be useful to humans in specific application contexts---another dimension of explanation quality that is distinct from their mathematical properties~\cite{10.1145/3411764.3445717, 10.1145/3397481.3450650}. Finally, many feature attribution methods are sensitive to one's choice of baseline~\cite{haug2021baselines, mamalakis2022carefully,sturmfels2020visualizing}

In this paper, we focus on measuring faithfulness~\cite{NEURIPS2018_3e9f0fc9}. The basic concept behind faithfulness is that feature attributions output by the XAI method should reproduce the actual influences of the features in the model. Faithfulness can be measured through various quantitative metrics, such as Prediction Gap on Important feature perturbation (PGI) and Prediction Gap on Unimportant feature perturbation (PGU) (first defined in~\cite{Dai_2022}, but given these names in \cite{Chirag_2022}).

\subsection{Post-hoc explainability benchmarking}

In an attempt to facilitate easier comparison of explanation methods, recent papers have introduced XAI benchmarks for feature attribution-based post-hoc explanations~\cite{Dai_2022, belaid2022need, Chirag_2022}. We add to this literature by introducing two additional metrics: ablation and topological data analysis. We also compare them to other faithfulness metrics.

\section{Methodology}

\begin{figure}[h]
    \centering
  \includegraphics[width=0.8\textwidth, ,bb=0 0 100 100, trim={0 0pt 0 25pt},clip]{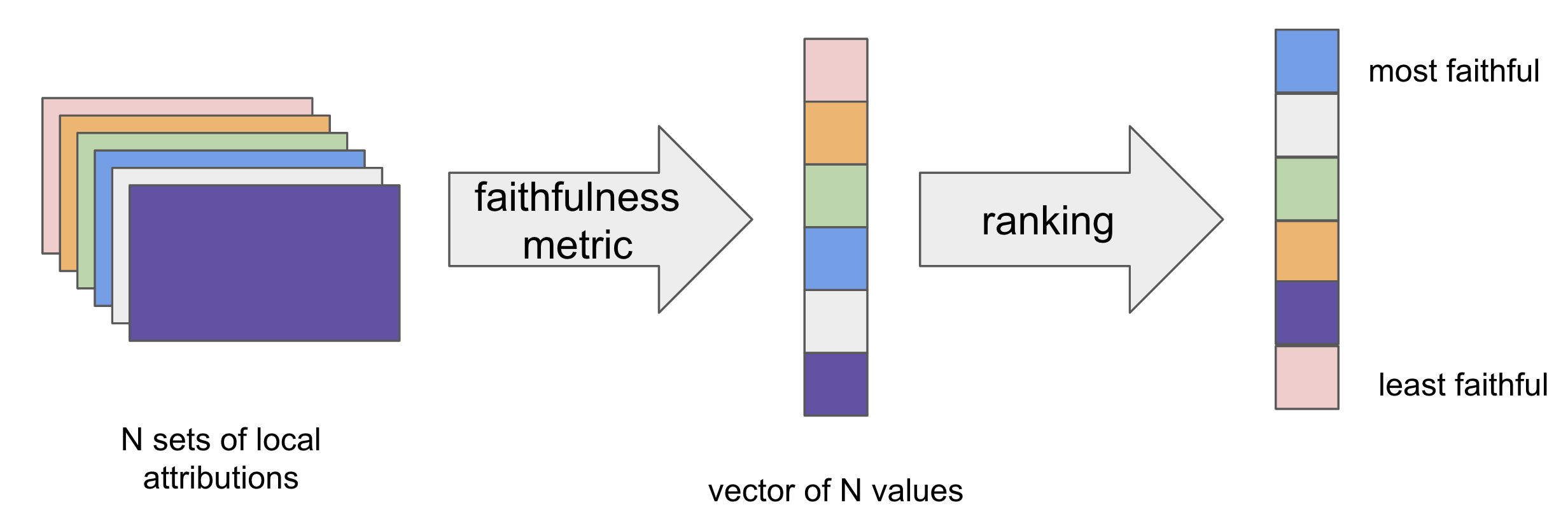}
  \caption{Cartoon of the data flow / process.  If the faithfulness metric was valid, this would allow a practitioner to choose the best explanations for their model.}
  \label{fig:diagram}
\end{figure}

The framework for the study is to use models built on commonly used empirical data with a variety of local explanations generated for each model-data pair.  Using those explanations along with a scalar metric intended to measure faithfulness, we generate a rank ordered list that in theory should give guidance on choosing the most faithful set of explanations.  We compare these rankings in~\ref{section:experiments}.

\subsection{Data and models}
We use one synthetic and four open source datasets~\cite{Dua:2019} whose characteristics are shown in Table~\ref{tab:data}.  For each dataset, we built a fixed size dense neural network with three layers and a single layer neural network as our linear model. See Table \ref{tab:data} in the appendix for details on the open-source datasets.

\subsection{Explanations}
For both the linear and non-linear models, we used three methods from the Captum package~\cite{captum}: Deep SHAP, KernelSHAP, and integrated gradients \cite{10.1145/3351095.3372850, sundararajan2017axiomatic}. However, specifying methods alone is not sufficient~\cite{haug2021baselines}. Each of these methods produces different outputs depending on the choice of baseline: for example, the popular open source \texttt{shap} package uses the average of all predictions as its default baseline \cite{10.1145/3351095.3372850}.  We also include a set of random explanations, uniformly sampled from [0,1] to provide a lower limit of faithfulness. For linear models, we include ground truth explanations of the form $x_i c_i$ where $x_i$ is the $i$th input feature and $c_i$ is the learned coefficient. 

For each explanation method, we supplied four baselines: \textit{constant median}, a variant of the constant baseline that uses  the median of each feature; \textit{training}, also known as the expectation baseline, a sub-sample of the training data \cite{lundberg2017unified};  \textit{opposite class}~\cite{albini2021counterfactual}, which selects $k$ samples that belong to the opposite 
 class from the the chosen sample; and \textit{nearest neighbor}~\cite{albini2021counterfactual}, which selects $k$ samples that are closest to the chosen sample. For this study the value of $k$ is fixed at five.

 Evaluation of each method-baseline pair is repeated three times to account for the possibility of stochastic behavior.  For the dense networks, we have a total of 39 tables of local explanations (3 methods with 4 baselines each and the random explanations, all with 3 repeats).  For the linear models, we have a set of 42 tables of explanations (13 in common with the nonlinear models with the addition of the ground truth explanations, all with 3 repeats).

\subsection{Metrics of faithfulness}

We sample existing metrics of faithfulness from the literature and open source repositories, and add two novel methods: ablation~\cite{hameed2022based} and topological data analysis~\cite{xenopoulos22}. 



\label{section:PGI}
PGI is calculated by measuring the change in a model's prediction when the top $k$ most important features, as determined by an explanation method, are perturbed.  The intuition behind the metric is that a model's output should change most dramatically when the most important features are perturbed. A higher PGI indicates a more faithful explanation~\cite{Dai_2022}. PGI is defined as follows:

\begin{equation}
\hat{\mathcal{M}}_{\text{PGI}}(\mathbf{x}, f)=\frac{1}{m} \sum_{j=1}^m\left[\left|f(\mathbf{x})-f\left(\tilde{\mathbf{x}}_j\right)\right|\right]
\end{equation}
Where $\mathbf{x}$ is the original sample, $\tilde{\mathbf{x}}$ is the same sample with the top $k$ features perturbed, and $f$ is a model which outputs a value from 0 to 1. The average PGI is computed over $m$ runs of a stochastic perturbation process.  We use a Gaussian perturbation drawing samples from $\mathcal{N}(0,\,0.1)$ for continuous variables, and a ``flip percentage'' of 0.3 for discrete variables unless otherwise mentioned.


\textbf{Ablation} is another perturbation-based method.  This procedure is frequently used to assess the importance of input variables on a model's performance. By perturbing the input variables in rank order of importance as determined from local attributions, one can assess the quality of their rank ordering. Essentially, the goal is to assess the sensitivity of the model's performance as gauged by the local explanations. Perturbing important variables should correlate with larger decreases in measures of model capability than perturbing less important features.  Details on ablation studies can be found in~\cite{hameed2022based}. For ablation, we use area under the ablation curve (ABC) as the scalar faithfulness metric. 


\textbf{Bottleneck distance (BND)} is a similarity metric to compare two manifolds,  with origins in topological data analysis (TDA) and persistent homology. It does not rely on perturbations---a characteristic which makes its possible use appealing. Instead, this methodology treats the point cloud of explanations as a manifold.  Through the use of the mapper algorithm~\cite{singh2007topological} with a specified filter function (here we use the model predictions), the high-dimensional manifold is projected down to a two dimensional network representation called the Reeb graph.  The similarity of two mapper networks can be compared by the bottleneck distance.  Even though the technical underpinning and units of the local explanations may differ, the smaller the distance between their respective manifolds, the more they are similar~\cite{xenopoulos22}.  For each explanation method, we use its average bottleneck distance to all other methods as the metric of explanation quality. Details of the TDA process can be found in~\ref{supplementary:tda_process}.


\subsection{Ranking explanations}
For each dataset, with its corresponding set of explanations and controls, we calculate each of the metrics and rank order their values. From those rankings, we use rank correlations to measure their agreement.  In an ideal world, where the metrics can consistently assess faithfulness,  the resulting correlation would be one. Detailed plots of the rankings can be seen in Section~\ref{subsection:slope_charts_all_nn}.  Summary heatmaps of the correlations are shown in Figure~\ref{fig:correlation_heatmaps}.  


\section{Experiments}
\label{section:experiments}

\subsection{Ground truth ranking on synthetic dataset}
\begin{figure}[h!]
    \centering
    \begin{subfigure}{.28\textwidth}
        \centering
        \includegraphics[width=\textwidth,bb=0 0 100 100,trim={0 5pt 0 0},clip]{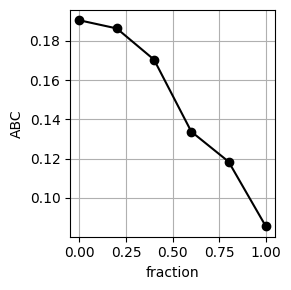}
        \label{fig:ground_truth_ranking_abc}
    \end{subfigure}
    ~ 
    \begin{subfigure}{.29\textwidth}
        \centering
        \includegraphics[width=\textwidth,bb=0 0 100 100,trim={0 5pt 0 0},clip]{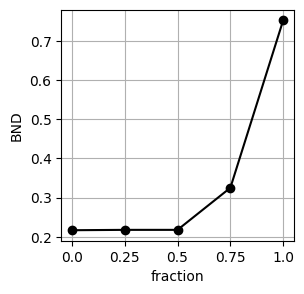}
        \label{fig:ground_truth_ranking_BND}
    \end{subfigure}
    ~
    \begin{subfigure}{.31\textwidth}
        \centering
        \includegraphics[width=\textwidth, bb=0 0 100 100,trim={0 5pt 0 0},clip]{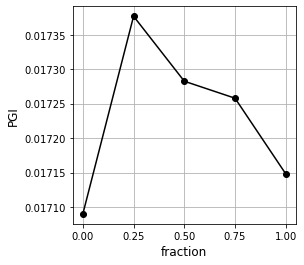}
        \label{fig:ground_truth_ranking_PGI}    
    \end{subfigure}
    \caption{Comparison of metrics (ABC, BND, and PGI respectively) on a set of explanations with a designed ranking.}
    \label{fig:ground_truth_metrics}
\end{figure}

The first experiment is to create a set of explanations with a controlled change in faithfulness.  Using the synthetic dataset, its logistic regression model, and ground truth explanations from the logits, we permute a portion of the rows of the explanation - with fractions varying from 0 to 1.  The explanations with more permuted rows are fundamentally less faithful since the explanations are increasingly misaligned with the inputs and their predictions.  The expected outcome is to have metrics that reflect this change in a strictly monotonic manner.  The results of the experiment are shown in Figure~\ref{fig:ground_truth_metrics}.

Of the three metrics, only the ABC metric matches our expectation.  The BND metric is the next closest to expectations---but it is not \textit{strictly} monotonic for fractions up to 0.5; while the PGI metric is not monotonic.  For details for the behavior of TDA for this experiment, detailing its lack of differentiation from 0 to 0.5, see Section~\ref{section:ground_truth_details}.  See Section~\ref{section:perturbation_considerations} for an analysis of the main effects of parameter choices for PGI.

\subsection{Experiments on non-synthetic datasets}

\begin{figure}[ht]
    \centering
    \begin{subfigure}{0.99\textwidth}
        \centering
        \includegraphics[width=\textwidth,bb=0 0 1200 100]{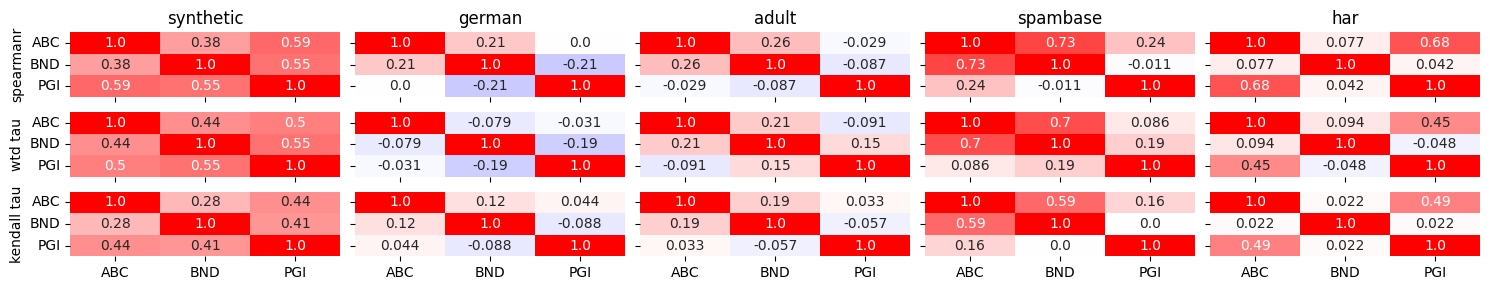}
        \caption{Logistic regression models.}
        \label{fig:heatmaps_logistic}
    \end{subfigure}
     
    \begin{subfigure}{0.99\textwidth}
        \centering
        \includegraphics[width=\textwidth,bb=0 0 1200 215]{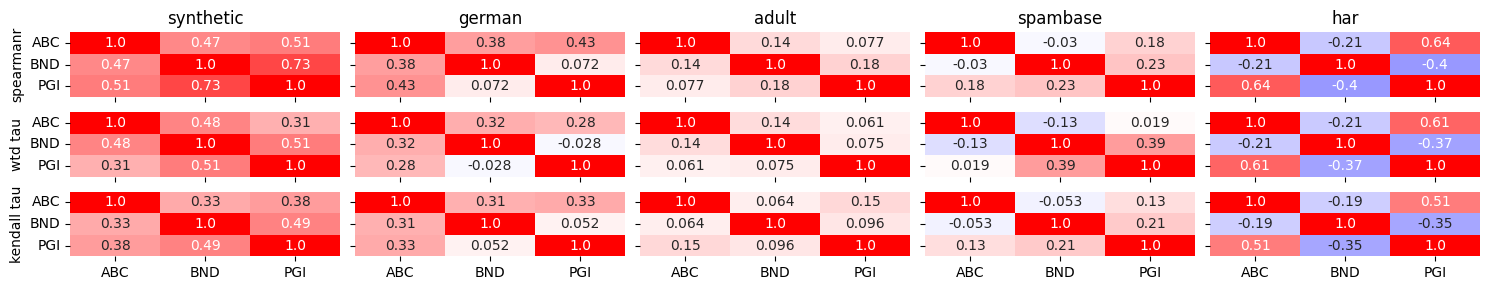}
        \caption{Three layered dense network models.}
        \label{fig:heatmaps_dense}
    \end{subfigure}
   \caption{Rank correlation heatmaps of the faithfulness metrics for each dataset in the study. The numbers are averaged across all combinations of baselines and explanation methods.}
   \label{fig:correlation_heatmaps}
\end{figure}

Figure~\ref{fig:correlation_heatmaps} shows the correlation of faithfulness metrics for each of the datasets in this study for logistic regression models (top) and dense networks (bottom) using Spearman's $\rho$ (top row), weighted Kendall-$\tau$ (middle row), and Kendall's $\tau$ (bottom row).   Using a logistic regression model, with the addition of the ``ground truth'' explanations, provides the opportunity to rank the XAI-generated explanations alongside ground truth and random explanations, with the expectation that the top ranked explanations will be the ground truth, and the worst ranked will be the random explanations.  For the dense networks the only known quantity is that the random explanations should be ranked the worst.  The correlations are insensitive to the choice of correlation metric.


\begin{figure}[h]
    \centering
    \begin{subfigure}{.48\textwidth}
        \centering
        \includegraphics[width=\textwidth,bb=0 0 100 100,trim={0 0 0 40pt},clip]{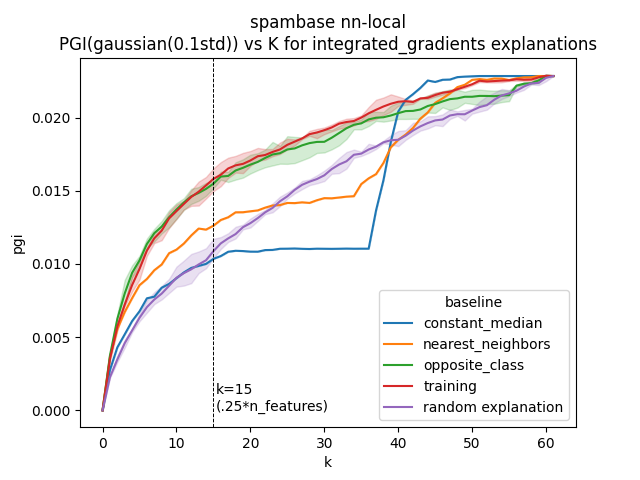}
        \caption{Gaussian perturbation}
        \label{fig:spambase_gaussian}
    \end{subfigure}
    ~ 
    \begin{subfigure}{.48\textwidth}
        \centering
        \includegraphics[width=\textwidth,bb=0 0 100 100,trim={0 0 0 55pt},clip]{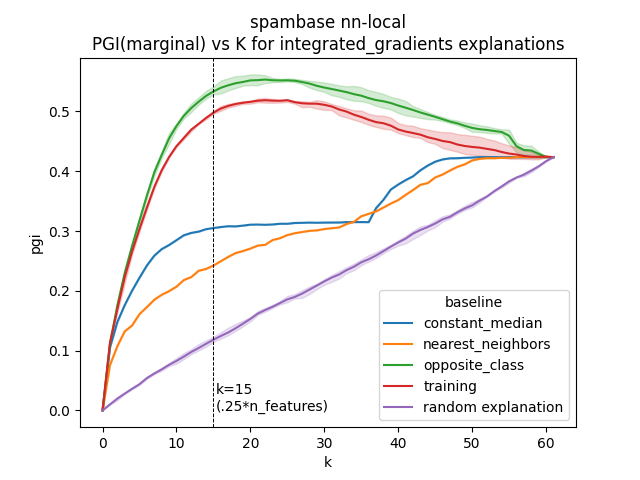}
        \caption{Marginal perturbation}
        \label{fig:spambase_marginal}
    \end{subfigure}
   \caption{Sensitivity study on choice of $k$ and perturbation for PGI on spambase with a dense network, with integrated gradients as the explanation method.}
   \label{fig:sensitivity}
\end{figure}



\subsection{Considerations for perturbation based metrics}
\label{section:perturbation_considerations}
\begin{figure}[h]
    \centering
  \includegraphics[width=0.75\textwidth, bb=0 0 100 100,trim={0 0 0 0.3cm},clip=True]{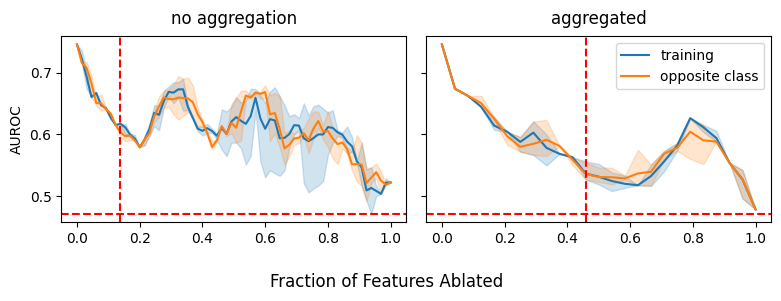}
  \caption{Sensitivity of ablation curves for the training and opposite class baselines. The vertical line depicts the best average rank of a random feature.}
  \label{fig:sensitivity_categorical}
\end{figure}

PGI and Ablation are both perturbation-based methods and share similar drawbacks~\cite{hooker2021extrapolation}.   Nonetheless, we see discrepancies between them in the correlation heatmaps in Figure~\ref{fig:correlation_heatmaps}.  While perturbation methods for explainability have been widely studied~\cite{covert2021}, that is not the case for faithfulness metrics.  Perturbation-based metrics of explanation faithfulness can vary widely depending on their numerous configuration parameters. Here we mention three:
\begin{enumerate}
    \item Choice of perturbation method: This is known to have a large, and biasing impact, on these types of ``permute and predict'' metrics~\cite{hooker2021extrapolation}.  In Figure~\ref{fig:sensitivity}, we contrast PGI calculated with a Gaussian perturbation (left) and with a marginal perturbation (right). For a fixed value of $k$ (shown with a vertical dashed line), higher values of PGI indicate higher faithfulness.  It is evident that the ranking of the baselines between the two figures are dissimilar.  For this instance, the Gaussian perturbation appears to perform more poorly, as it ranks the random explanation ahead of the constant median baseline.
    
    \item Choice of top $k$: how to select $k$ is not theoretically motivated.  We see in Figure~\ref{fig:sensitivity} that choice of k can impact the faithfulness rankings of a set of explanations based on PGI. Wherever the lines cross, PGI rankings change. In the Gaussian perturbation plot, three of the five baselines achieve the highest PGI rank at different values of k. At k$<=$11, the opposite class baseline has the highest PGI. Then, training baseline is the top ranked for 12$<=$k$<=$41,  until finally constant median takes over.
    
    \item Treatment of discrete features: this encompasses two issues: i) it is difficult to fairly balance the strength of Gaussian noise on continuous variables and the flip percentage on discrete variables; and ii) the choice of treating them as one-hot-encoded, instead of performing an aggregation and reverting the input features back to a nominal label-encoded state.  Figure~\ref{fig:sensitivity_categorical} shows the difference in calculated ablation curves on both one-hot-encoded and aggregated categorical features.  If the rank ordering of features is consistent, it is expected that the ablation curve would be monotonically decreasing---which is clearly not the case for no aggregations. For the aggregated plot, the ablation curves are closer to the ideal, with significant increases in AUROC occurring only after the random feature sanity check.
\end{enumerate}

\subsection{Observations on the use of TDA}
At the heart of TDA is the use of the bottleneck distance (BND), and choosing the set of explanations that have the lowest distance to the other candidates.  However, the distance is a similarity metric - it does not have a notion of ``good'' or ``faithful'' baked into it.  This can be limiting, as one cannot directly compare two items - the sum of the distances for each candidate would be identical.   In a larger collection of methods, if the candidates consisted of a majority of explanations of low quality and a minority of high quality, the selected candidate will most likely be chosen from the low quality pool. This should influence the construction of candidates for assessment.  The current process of selecting one resolution as a result of its performance on all candidates for a dataset seems to have diminished its sensitivity, which can be seen in its robustness to permuting rows in Figure~\ref{fig:ground_truth_metrics}, and for the selection of candidates for the synthetic dataset in Figure~\ref{fig:tda_choose_candidate}




\section{Conclusions}
We have focused on using XAI methods on a range of public empirical datasets, using the tabular data to create classification models---a typical use case in academia and industry. The goal was to find a set of explanations that was deemed to be most faithful.

Across the experiments, the ranked correlations reveal little consensus on the notion of faithfulness in the explanations.  This in turn would leave an end user without the required tools to know if they had successfully chosen the right set of explanation method and baseline.  

This gap in utility should be a wake-up call to the XAI community. Future work can compare the plethora of additional measures of faithfulness \cite{JMLR:v24:22-0142,Chirag_2022} to see if they also disagree.




%


\setcitestyle{numbers}

\bibliographystyle{plain}
\bibliography{07_main} 

\appendix
\section{Supplementary Material}
\label{section:supp}


%
%
\subsection{Data set information}
Table~\ref{tab:data} shows details of the synthetic and empirical data sets used in the experiments.

\begin{table}[h]
\centering
    \begin{tabular}{lrrrr}
    \toprule[1pt]

    Dataset        & Samples  & Num  & Cat & OHE \\ 
    \hline
    synthetic      &   1,000  & 24   & 0   & 0  \\
    adult          &  48,842  & 6    &  8  & 100 \\ 
    German credit  &   1,000  & 7    & 13  & 54  \\
    har            &  10,299  & 561  & 0   & 0   \\
    spambase       &   4,601  & 57    & 0  &  0 \\
    \hline
    \end{tabular}
    \vspace{0.1cm}
\caption{Summary of datasets, detailing number of samples, numerical (Num), categorical (Cat), and one hot encoded (OHE) features.}
\label{tab:data}
\vspace{-0.6cm}
\end{table}


\subsection{TDA process - hyperparameters, networks, and persistence diagrams}
\label{supplementary:tda_process}
\begin{figure}[h]
    \centering
    \includegraphics[width=\textwidth, bb=0 0 100 100,trim={0 2.5cm 0 2.5cm},clip=True]{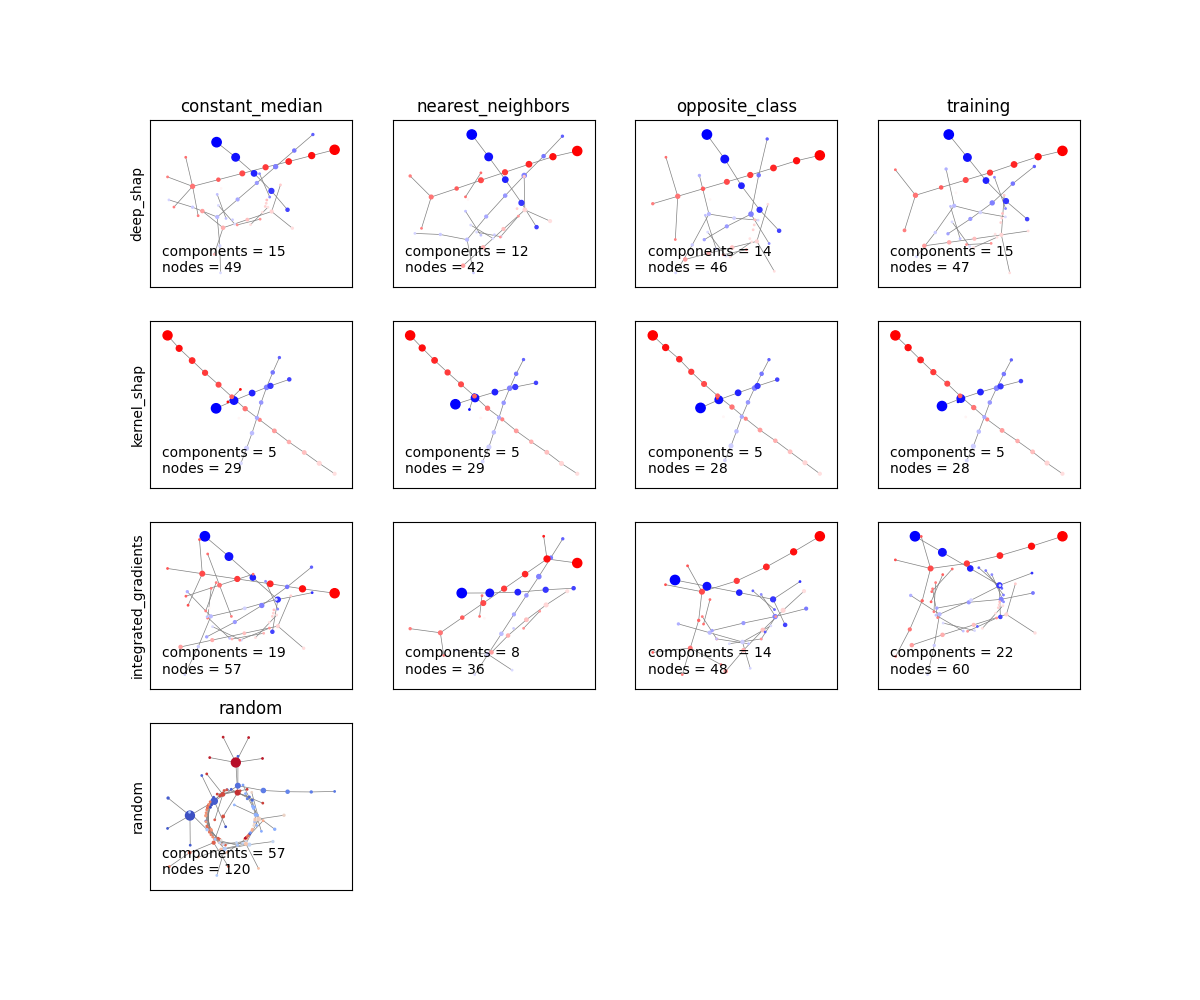}
    \caption{The network representation of a mapper object for each method and baseline for the synthetic dataset. }    
    \label{fig:tda_synthetic_mappers}
\end{figure}

In practice, a topological manifold is represented by a point cloud.  A mapper object is constructed from five things:
\begin{itemize}
    \item A manifold as represented by a point cloud - the explanations.
    \item A ``lens function`` that provides scalar valued labels to every point in the cloud.  Here we use the model predictions.
    \item A value for resolution. The resolution sets the number of slices that divide contiguous ranges of the lens function.  To choose the value of resolution, we conduct a grid search over a range of values from six to thirty by twos. Each value of resolution is calculated over 30 bootstrap samples and we collect the 95th percentile confidence value from the bootstrap bottle neck distances as  our figure of merit, called the \textit{stability}.  For a dataset,  we collect a table of stability values from all methods, baselines and repeats, for each candidate resolution.  The resolution with the lowest total sum of stability is selected.  
    \item A clustering algorithm that operates on the points in a slice.  This clustering creates the nodes of the network.  The majority of the theoretical work in TDA has used agglomerative clustering - which we also adopt.  The GUDHI~\cite{gudhi:urm} library offers utility functions to estimate its distance parameter, which eliminates a potential search for that hyperparameter.
    \item A value for the gain.  The gain specifies the overlap of neighboring slices. Theoretical analysis shows a valid range for gain is from 0.3 to 0.5~\cite{carriere2018statistical}.  We choose a gain value of 0.4 for all results shown in this study.  Points that exist in the overlap region create edges between nodes.  
\end{itemize}

\begin{figure}[!ht]
    \centering
    \includegraphics[width=\textwidth, bb=0 0 100 100,trim={0 2.5cm 0 2.5cm},clip=True]{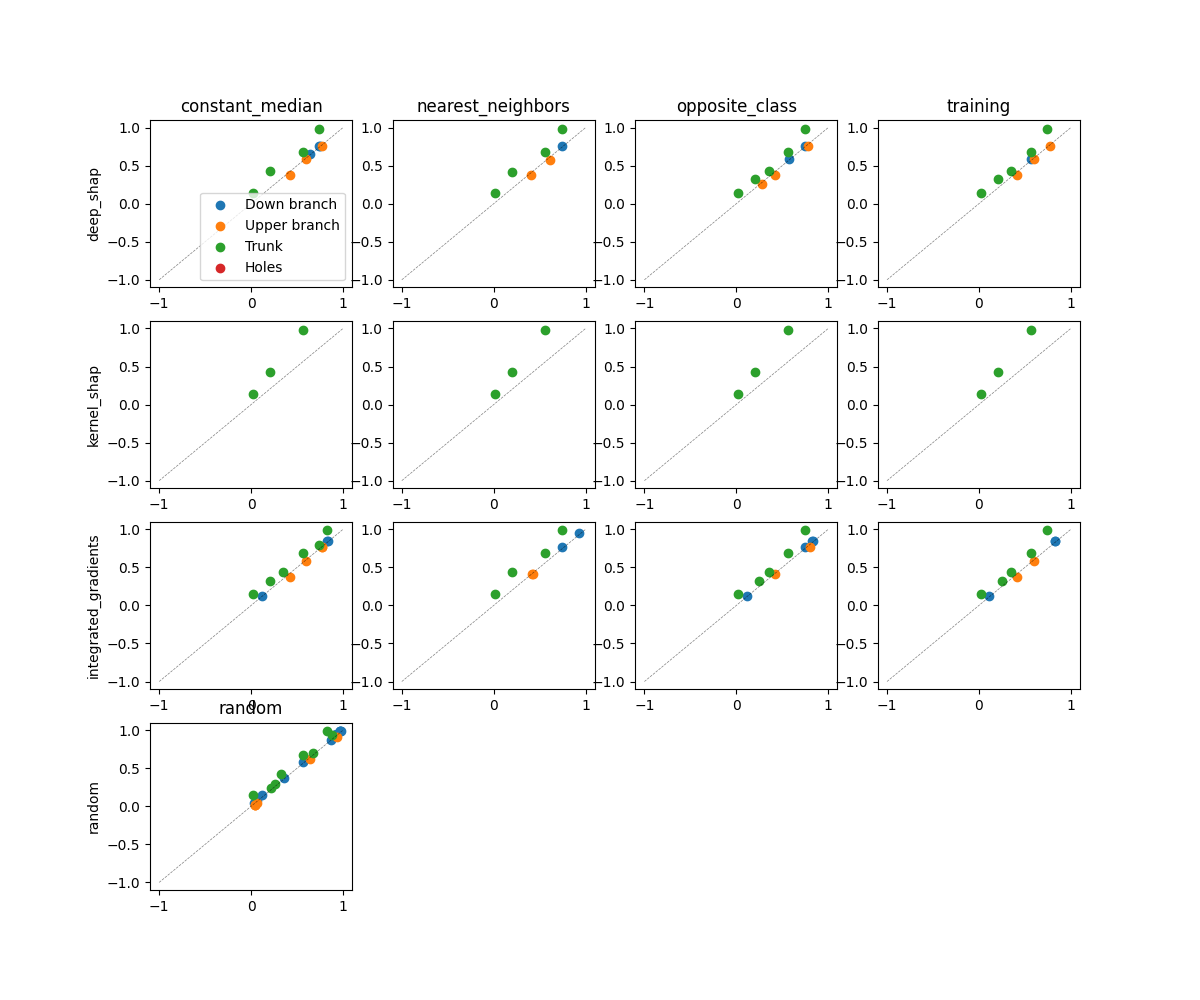}
    \caption{Persistence diagrams corresponding to the networks shown in Figure~\ref{fig:tda_synthetic_mappers}.}
    \label{fig:tda_synthetic_pds}
\end{figure}

The mapper networks for one repeat are shown in Figure~\ref{fig:tda_synthetic_mappers}.  For each network, a persistence diagram is created,  shown in Figure~\ref{fig:tda_synthetic_pds}. Each point on the diagram represents a single simple connected component, or fork, or hole that exists over a range of values for the filter function (in this case the model predictions).  The x-axis specifies when a topological feature is ``born'' (when it first appears as measured by the filter function) while the y-axis specifies when a topological feature ``dies'' (the final value of the filter function for the last node).

\begin{figure}[ht]
    \centering
    \begin{subfigure}{.48\textwidth}
        \centering
        \includegraphics[width=\textwidth,bb=0 0 900 400]{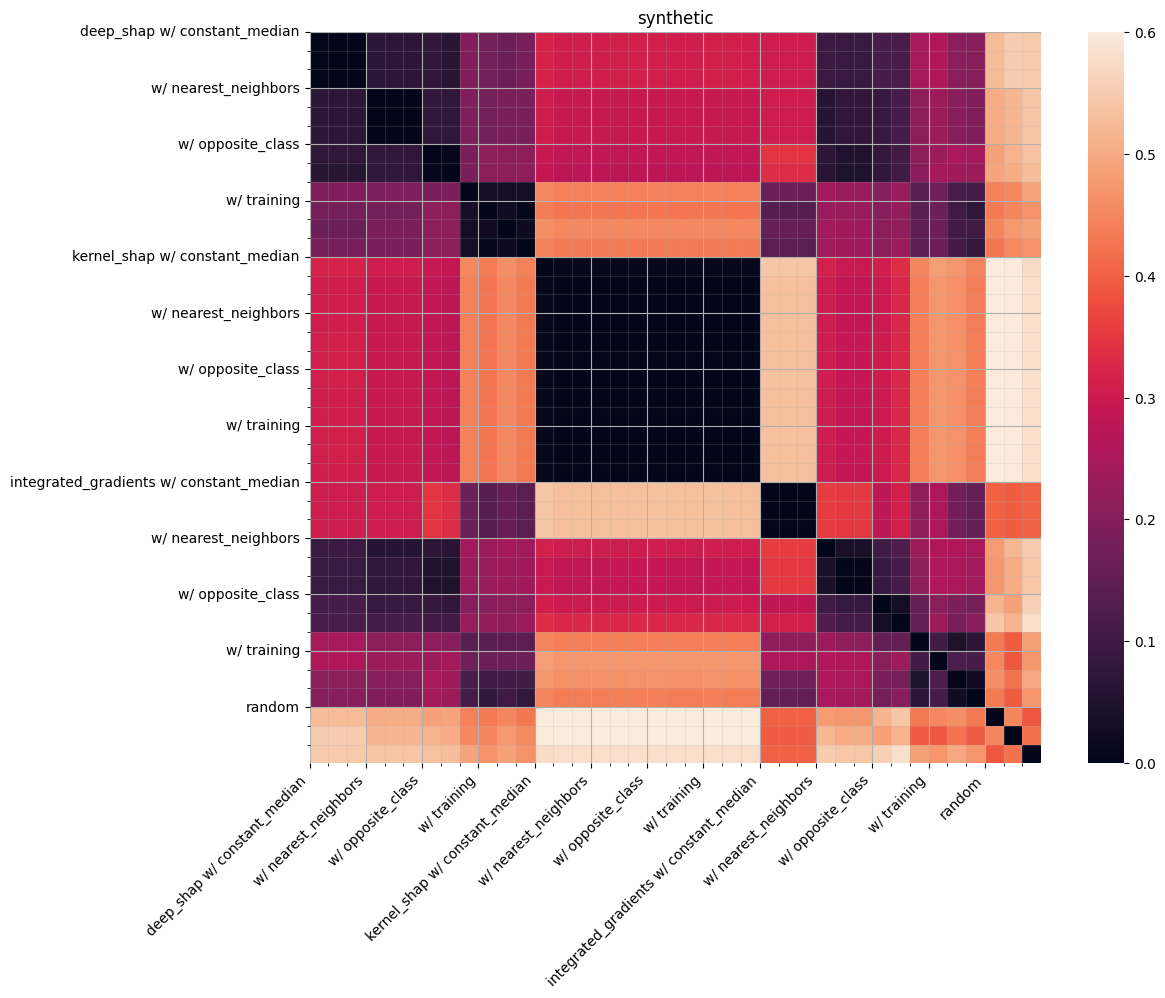}
    \end{subfigure}
    ~ 
    \begin{subfigure}{.48\textwidth}
        \centering
        \includegraphics[width=\textwidth, bb=0 0 600 400]{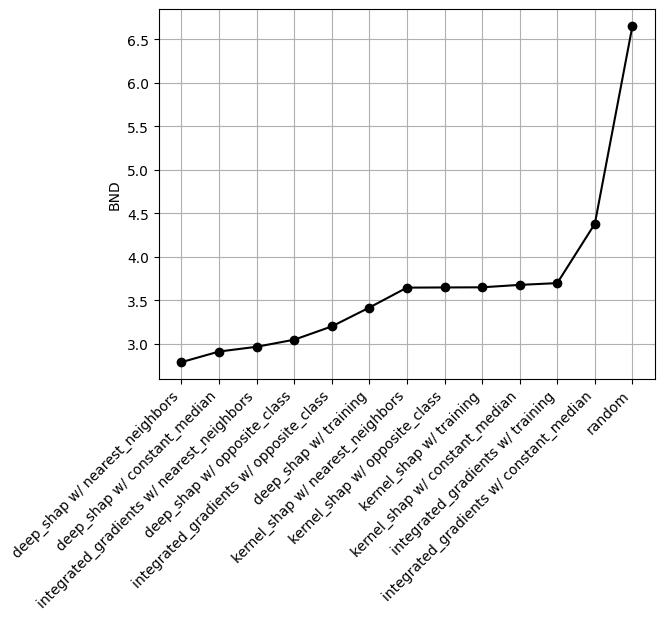}
    \end{subfigure}
    \caption{Heatmap of bottleneck distance, baseline and repeat (left) and its sum (right) for all candidates.  The candidate with the lowest sum is considered the best.}
    \label{fig:tda_choose_candidate}
\end{figure}

The \textit{bottleneck distance} (BND) is calculated by matching the entities on two persistence diagrams and measuring the euclidean distance between the points on the persistence diagram.  A heatmap showing these distances is shown in the left of Figure~\ref{fig:tda_choose_candidate}.  The candidate is chosen to be the one with the lowest row-wise sum of the bottleneck distances, as seen in the right of Figure~\ref{fig:tda_choose_candidate}.

%
%
\subsection{Ground truth ranking details for BND}
\label{section:ground_truth_details}
Here we show how low values of permutation (0, 0.25 and 0.5) ``look the same'' to the BND metric - with little change in the mappers, resulting in no change to the bottleneck distance.  With permutations of 0.25 and 0.5, the mappers show a disconnected cluster, which does not persist, and does not amount to a significant differentiation.  It's not until the permutation reaches 0.75 that a new persistent feature (the upper branch) is detected, and more features continue to be added at 1.0 leading to a significant difference being found by BND.
\begin{figure}[ht!]
    \centering
    \begin{subfigure}{\textwidth}
        \centering
        \includegraphics[width=\textwidth,bb=0 0 1100 100]{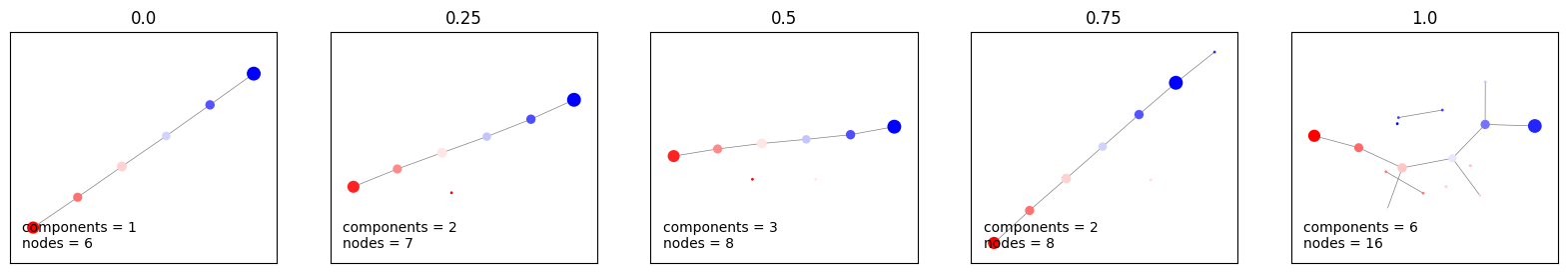}
        \label{fig:fig:gt_bnd_mappers}
    \end{subfigure}
     
    \begin{subfigure}{\textwidth}
        \centering
        \includegraphics[width=\textwidth,bb=0 0 1100 200]{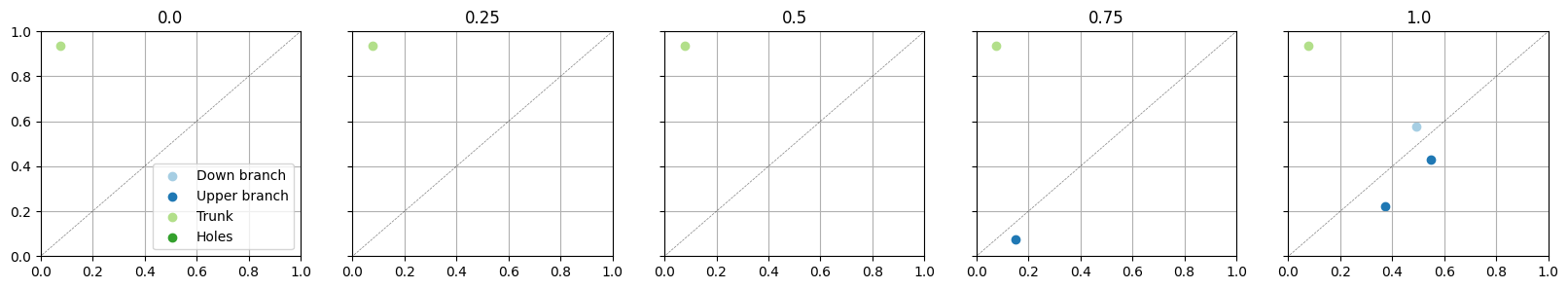}
        \label{fig:gt_bnd_pd}
    \end{subfigure}
   \caption{For the ground truth ranking experiment, plots of all mapper networks (top) and corresponding persistence diagrams (bottom).}
   \label{fig:BND_robustness}
\end{figure}

\subsection{Top three most faithful explanations}
While it is possible the rank correlation could over weight mismatches in lower ranked components, Table~\ref{ref:table_winners} shows the three most faithful explanations as chosen by each metric.  It is clear to see that there is no consistent agreement among the metrics.  However if taken in aggregate, the methods that appears most often is integrated gradients (18), followed by deep shap (17), and lastly by kernel shap (10).  For baselines, the picture is clearer - the one that appears most frequently is opposite class (24), followed by training (14) and tied in last place are constant median (4) and nearest neighbors(4).


\begin{table}[ht]
\centering
\renewcommand{\arraystretch}{1.05}
 \begin{tabular}{ccllc}
\toprule
dataset & metric & method & baseline & ranking  \\
\midrule
\multirow{2}{*}[-4.2em]{synthetic} & ABC & deep-shap & opposite-class & 0 \\
     & ABC & integrated-gradients & opposite-class & 1 \\
     & ABC & integrated-gradients & training & 2 \\
     \cline{2-5} \noalign{\vspace{0.5ex}}
     & BND & deep-shap & nearest-neighbors & 0 \\
     & BND & deep-shap & constant-median & 1 \\
     & BND & integrated-gradients & nearest-neighbors & 2 \\
     \cline{2-5} \noalign{\vspace{0.5ex}}
     & PGI & deep-shap & opposite-class & 0 \\
     & PGI & integrated-gradients & opposite-class & 1 \\
     & PGI & integrated-gradients & training & 2 \\
\cline{1-5}
\multirow{2}{*}[-4.2em]{german} & ABC & kernel-shap & training & 0 \\
     & ABC & kernel-shap & opposite-class & 1 \\
     & ABC & deep-shap & opposite-class & 2 \\
     \cline{2-5} \noalign{\vspace{0.5ex}}
     & BND & deep-shap & opposite-class & 0 \\
     & BND & integrated-gradients & nearest-neighbors & 1 \\
     & BND & kernel-shap & training & 2 \\
     \cline{2-5} \noalign{\vspace{0.5ex}}
     & PGI & kernel-shap & training & 0 \\
     & PGI & kernel-shap & opposite-class & 1 \\
     & PGI & deep-shap & opposite-class & 2 \\
\cline{1-5}
\multirow{2}{*}[-4.2em]{adult} & ABC & deep-shap & opposite-class & 0 \\
     & ABC & integrated-gradients & opposite-class & 1 \\
     & ABC & integrated-gradients & training & 2 \\
     \cline{2-5} \noalign{\vspace{0.5ex}}
     & BND & deep-shap & constant-median & 0 \\
     & BND & deep-shap & nearest-neighbors & 1 \\
     & BND & deep-shap & opposite-class & 2 \\
     \cline{2-5} \noalign{\vspace{0.5ex}}
     & PGI & deep-shap & opposite-class & 0 \\
     & PGI & integrated-gradients & opposite-class & 1 \\
     & PGI & integrated-gradients & training & 2 \\
\cline{1-5}
\multirow{2}{*}[-4.2em]{spambase} & ABC & integrated-gradients & opposite-class & 0 \\
     & ABC & deep-shap & opposite-class & 1 \\
     & ABC & kernel-shap & opposite-class & 2 \\
     \cline{2-5} \noalign{\vspace{0.5ex}}
     & BND & deep-shap & constant-median & 0 \\
     & BND & deep-shap & training & 1 \\
     & BND & integrated-gradients & opposite-class & 2 \\
     \cline{2-5} \noalign{\vspace{0.5ex}}
     & PGI & integrated-gradients & opposite-class & 0 \\
     & PGI & deep-shap & opposite-class & 1 \\
     & PGI & kernel-shap & opposite-class & 2 \\
\cline{1-5}
\multirow{2}{*}[-4.2em]{har} & ABC & integrated-gradients & opposite-class & 0 \\
     & ABC & integrated-gradients & training & 1 \\
     & ABC & deep-shap & opposite-class & 2 \\
     \cline{2-5} \noalign{\vspace{0.5ex}}
     & BND & kernel-shap & opposite-class & 0 \\
     & BND & kernel-shap & training & 1 \\
     & BND & kernel-shap & constant-median & 2 \\
     \cline{2-5} \noalign{\vspace{0.5ex}}
     & PGI & integrated-gradients & opposite-class & 0 \\
     & PGI & integrated-gradients & training & 1 \\
     & PGI & deep-shap & opposite-class & 2 \\
\cline{1-5}
\bottomrule
\end{tabular}
\caption{Top three choices for each dataset broken down by metric. }
\label{ref:table_winners}
\end{table}

\subsection{Slope charts for dense networks}
\label{subsection:slope_charts_all_nn}

\begin{figure}[h!]
    \centering
    \textbf{Synthetic}\par\medskip 
    \begin{subfigure}{.3\textwidth}
        \centering
        \includegraphics[width=\textwidth, bb=0 0 250 100]{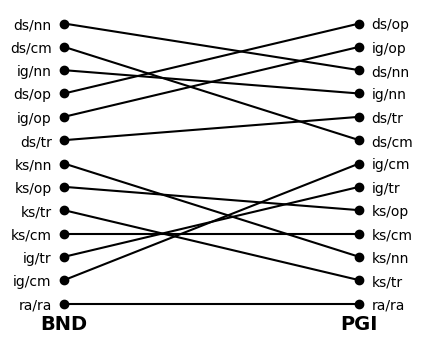}
        \caption{Ranking of BND and PGI}
        \label{fig:slope_left_synthetic}
    \end{subfigure}
    ~ 
    \begin{subfigure}{.3\textwidth}
        \centering
        \includegraphics[width=\textwidth, bb=0 0 250 100]{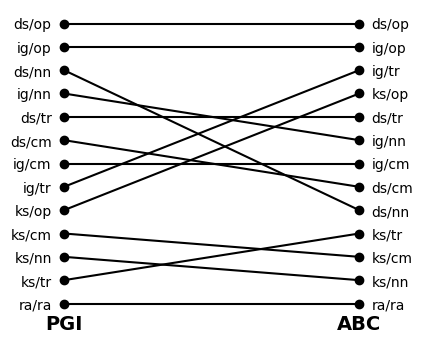}
        \caption{Ranking of PGI and ABC}
        \label{fig:slope_middle_synthetic}
    \end{subfigure}
    ~
    \begin{subfigure}{.3\textwidth}
        \centering
        \includegraphics[width=\textwidth, bb=0 0 250 100]{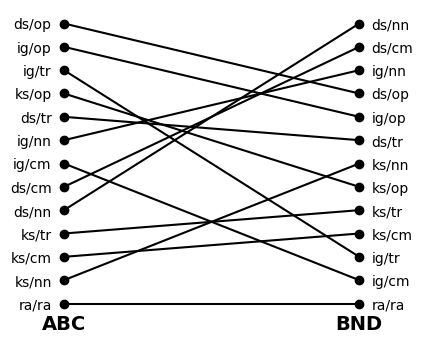}
        \caption{Ranking of ABC and BND}
        \label{fig:slope_right_synthetic}
    \end{subfigure}
    \caption{Slope charts for synthetic dataset}
    \label{fig:slope-chart_synthetic}
\end{figure}

\begin{figure}[h!]
    \centering
    \textbf{German credit}\par\medskip 
    \begin{subfigure}{.3\textwidth}
        \centering
        \includegraphics[width=\textwidth,bb=0 0 250 100]{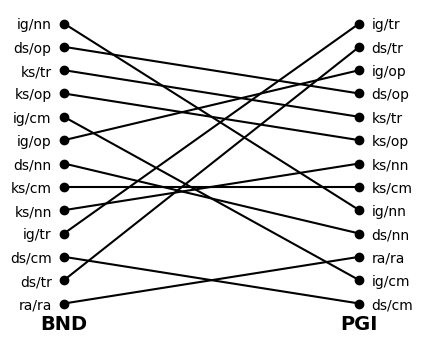}
        \caption{Ranking of BND and PGI}
        \label{fig:slope_left_german}
    \end{subfigure}
    ~ 
    \begin{subfigure}{.3\textwidth}
        \centering
        \includegraphics[width=\textwidth,bb=0 0 250 100]{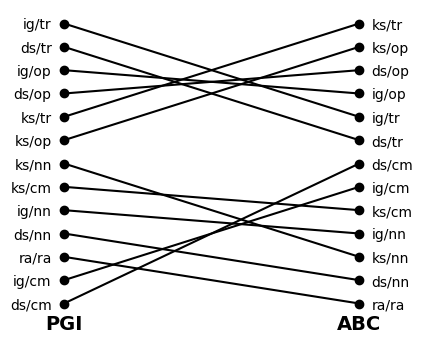}
        \caption{Ranking of PGI and ABC}
        \label{fig:slope_middle_german}
    \end{subfigure}
    ~
    \begin{subfigure}{.3\textwidth}
        \centering
        \includegraphics[width=\textwidth,bb=0 0 250 100]{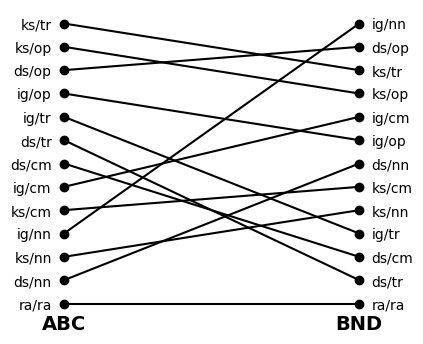}
        \caption{Ranking of ABC and BND}
        \label{fig:slope_right_german}
        \end{subfigure}
    \caption{Slope charts for German Credit dataset}
    \label{fig:slope-chart_german}
\end{figure}

\begin{figure}[h!]
    \centering
    \textbf{Spambase}\par\medskip 
    \begin{subfigure}{.3\textwidth}
    \centering
    \includegraphics[width=\textwidth,bb=0 0 250 100]{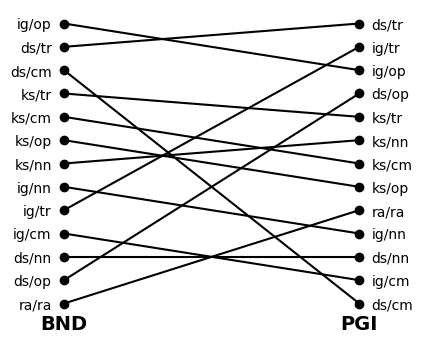}
    \caption{Ranking of BND and PGI}
    \label{fig:slope_left_spambase}
    \end{subfigure}
    ~ 
    \begin{subfigure}{.3\textwidth}
    \centering
    \includegraphics[width=\textwidth,bb=0 0 250 100,]{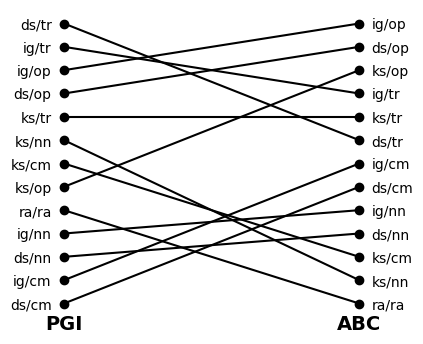}
    \caption{Ranking of PGI and ABC}
    \label{fig:slope_middle_spambase}
    \end{subfigure}
    ~
    \begin{subfigure}{.3\textwidth}
    \centering
    \includegraphics[width=\textwidth,bb=0 0 250 100,]{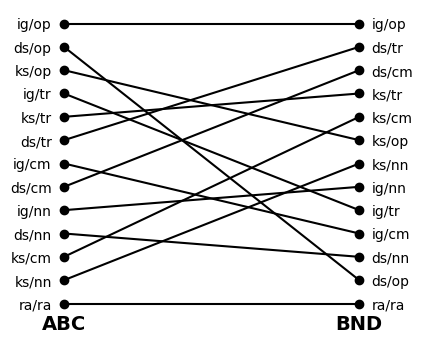}
    \caption{Ranking of ABC and BND}
    \label{fig:slope_right_spambase}
    \end{subfigure}
    \caption{Slope charts for spambase dataset}
    \label{fig:slope-chart_spambase}
\end{figure}

\begin{figure}[h!]
    \centering
    \textbf{Human Activity Recognition (HAR)}\par\medskip 
    \begin{subfigure}{.3\textwidth}
    \centering
    \includegraphics[width=\textwidth,bb=0 0 250 100]{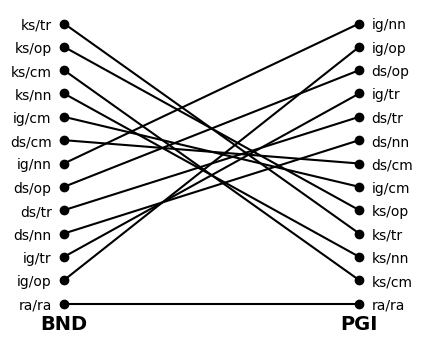}
    \caption{Ranking of BND and PGI}
    \label{fig:slope_left_har}
    \end{subfigure}
    ~ 
    \begin{subfigure}{.3\textwidth}
    \centering
    \includegraphics[width=\textwidth,bb=0 0 250 100]{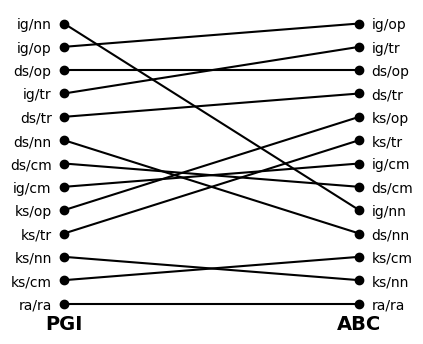}
    \caption{Ranking of PGI and ABC}
    \label{fig:slope_middle_har}
    \end{subfigure}
    ~
    \begin{subfigure}{.3\textwidth}
    \centering
    \includegraphics[width=\textwidth,bb=0 0 250 100]{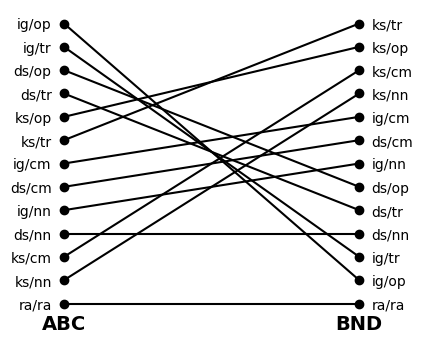}
    \caption{Ranking of ABC and BND}
    \label{fig:slope_right_har}
    \end{subfigure}
    \caption{Slope charts for human activity recognition dataset}
    \label{fig:slope-chart_har}
\end{figure}

The fundamental information being compared are the rankings of the explanations based on the faithfulness metrics.  To provide a more tangible sense of the mismatches, we have utilized slope charts (Figures \ref{fig:slope-chart_synthetic}, \ref{fig:slope-chart_german},  \ref{fig:slope-chart_spambase}, \ref{fig:slope-chart_har}) as means of interrogating the results.  Best ranked explanations are at the top, worst ranked at the bottom. When metrics place the explanations at a different rank, it can be  identified as a sloped line, with a steeper slope signifying a larger disagreement in rank.

\end{document}